\documentclass[conference]{IEEEtran}
\IEEEoverridecommandlockouts

\usepackage{cite}
\usepackage{enumitem}
\usepackage{amsmath,amssymb,amsfonts}
\usepackage{algorithmic}
\usepackage{graphicx}
\usepackage{textcomp}
\usepackage{xcolor}
\usepackage{forest}
\usetikzlibrary{trees,positioning,shapes,shadows,arrows.meta}
\usepackage{tikz}

\makeatletter
\newcommand{\linebreakand}{%
  \end{@IEEEauthorhalign}
  \hfill\mbox{}\par
  \mbox{}\hfill\begin{@IEEEauthorhalign}
}
\makeatother

\begin{document}

\title{From Word Vectors to Multimodal Embeddings: Techniques, Applications, and Future Directions For Large Language Models}

\author{
    \IEEEauthorblockN{
        Charles Zhang\textsuperscript{a, b},
        Benji Peng\textsuperscript{*, a, b},
        Xintian Sun\textsuperscript{c}, \\
        Qian Niu\textsuperscript{d}, 
        Junyu Liu\textsuperscript{d}, 
        Keyu Chen\textsuperscript{a},
        Ming Li\textsuperscript{a},
        Pohsun Feng\textsuperscript{e},
        Ziqian Bi\textsuperscript{f}, 
        Ming Liu\textsuperscript{f},\\
        Yichao Zhang\textsuperscript{g},
        Xinyuan Song\textsuperscript{h},
        Cheng Fei\textsuperscript{i},
        Caitlyn Heqi Yin\textsuperscript{j},
        Lawrence KQ Yan\textsuperscript{k},
        Tianyang Wang\textsuperscript{l},
        Hongyang He\textsuperscript{m},
    }
    \IEEEauthorblockA{
        \textsuperscript{a}Georgia Institute of Technology, USA
    }
    \IEEEauthorblockA{
        \textsuperscript{b}AppCubic, USA
    }
    \IEEEauthorblockA{
        \textsuperscript{c}Simon Fraser University, Canada
    }

    \IEEEauthorblockA{
        \textsuperscript{d}Kyoto University, Japan
    }
    \IEEEauthorblockA{
        \textsuperscript{e}National Taiwan Normal University, ROC
    }
    \IEEEauthorblockA{
        \textsuperscript{f}Purdue University, USA
    }
    \IEEEauthorblockA{
        \textsuperscript{g}The University of Texas at Dallas, USA
    }
    \IEEEauthorblockA{
        \textsuperscript{h}Emory University, USA
    }
    \IEEEauthorblockA{
        \textsuperscript{i}Cornell University, USA
    }
    \IEEEauthorblockA{
        \textsuperscript{j}University of Wisconsin-Madison, USA
    }
    \IEEEauthorblockA{
        \textsuperscript{k}The Hong Kong University of Science and Technology, PRC
    }
    \IEEEauthorblockA{
        \textsuperscript{l}University of Liverpool, UK
    }
    \IEEEauthorblockA{
        \textsuperscript{m}University of Warwick, UK
    }
    \IEEEauthorblockA{
        *Corresponding Email: benji@appcubic.com
    }
}

\maketitle


\begin{abstract}
Word embeddings and language models have transformed natural language processing (NLP) by facilitating the representation of linguistic elements in continuous vector spaces. This review visits foundational concepts such as the distributional hypothesis and contextual similarity, tracing the evolution from sparse representations like one-hot encoding to dense embeddings including Word2Vec, GloVe, and fastText. We examine both static and contextualized embeddings, underscoring advancements in models such as ELMo, BERT, and GPT and their adaptations for cross-lingual and personalized applications. The discussion extends to sentence and document embeddings, covering aggregation methods and generative topic models, along with the application of embeddings in multimodal domains, including vision, robotics, and cognitive science. Advanced topics such as model compression, interpretability, numerical encoding, and bias mitigation are analyzed, addressing both technical challenges and ethical implications. Additionally, we identify future research directions, emphasizing the need for scalable training techniques, enhanced interpretability, and robust grounding in non-textual modalities. By synthesizing current methodologies and emerging trends, this survey offers researchers and practitioners an in-depth resource to push the boundaries of embedding-based language models.
\end{abstract}

\section{Introduction}

Large Language Models (LLMs) have transformed natural language processing (NLP) by providing advanced tools for understanding and generating human language. At the core of these models are word embeddings—dense, continuous vector representations that capture semantic and syntactic relationships among words. By mapping words into high-dimensional spaces where semantically related words are situated near each other, embeddings support nuanced language interpretation and have become essential to NLP applications such as machine translation, sentiment analysis, and information retrieval. The progression from early one-hot encodings to more sophisticated embeddings like Word2Vec, GloVe, and fastText has markedly improved the accuracy and scalability of language models, enabling them to handle large volumes of textual data with high precision \cite{rifat_robust_2020, victor_language_2016, novotn_y_when_2022}.

Architectures like ELMo, BERT, and GPT employ deep neural networks to generate embeddings that reflect context-dependent meanings, addressing complexities such as polysemy and capturing long-range dependencies in language. These contextual representations not only enhance the accuracy and robustness of NLP systems but also support the integration of language with other modalities, including vision and robotics, enabling more sophisticated, interactive applications. However, despite these advancements, challenges persist concerning efficiency, interpretability, and ethical concerns within embedding models. High computational costs, the opaque nature of embedding spaces, and the risk of propagating biases from training data highlight areas for continued research \cite{peng2024jailbreaking, peng2024securing, li2024surveying}. This review examines the evolution of word embeddings, exploring foundational principles, varied methodologies, cross-modal applications, and key challenges.

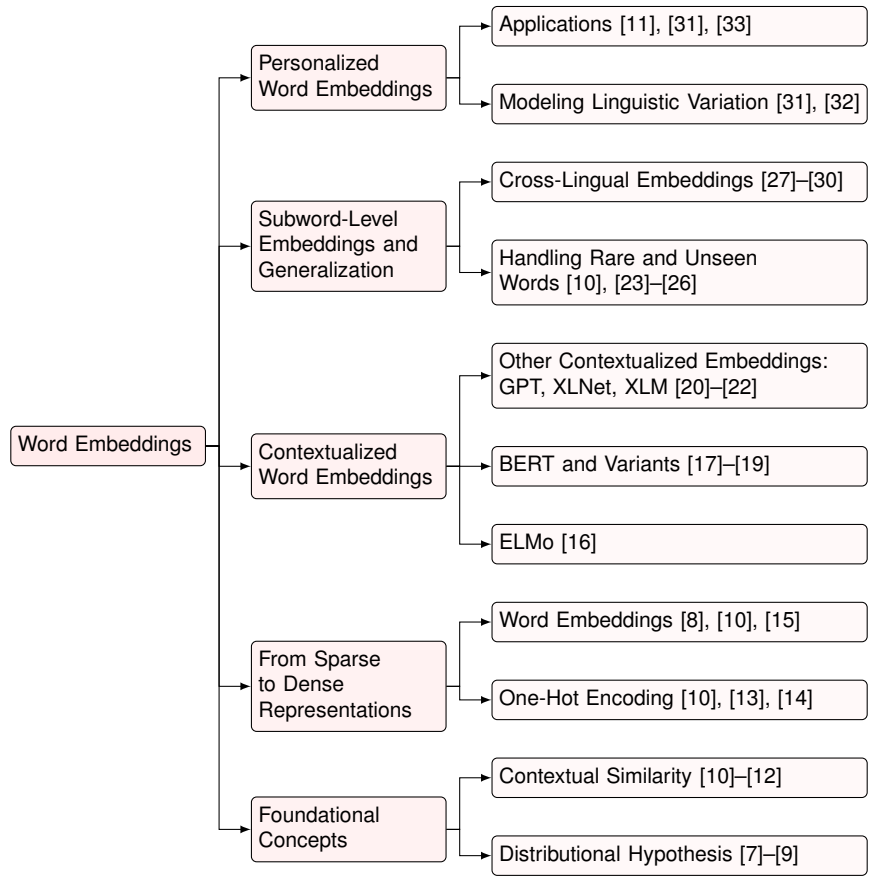
\begin{figure*}
    \centering
    
\tikzset{
    basic/.style  = {draw, align=left, font=\sffamily, rectangle},
    root/.style   = {basic, rounded corners=2pt, thin, fill=pink!30, text width=3cm, anchor=west},
    level1/.style = {basic, thin, rounded corners=2pt, fill=pink!20, text width=3cm},
    level2/.style = {basic, thin, rounded corners=2pt, fill=pink!10, text width=6cm},
    leaf/.style   = {basic, thin, fill=white!80!pink, text width=6cm},
    edge from parent/.style={draw=black, edge from parent fork right},
    level distance=2cm,
}

\begin{forest}
for tree={
    grow=east,
    scale=0.8, 
    growth parent anchor=west,
    parent anchor=east,
    child anchor=west,
    l sep=6mm, 
    s sep=5mm, 
    edge path={
        \noexpand\path[\forestoption{edge},->, >={latex}] 
        (!u.parent anchor) -- +(5pt,0pt) |- (.child anchor) 
        \forestoption{edge label};
    },
    align=left, 
}
[Word Embeddings, root,
    [Foundational \\ Concepts, level1,
        [Distributional Hypothesis \cite{lebret_word_2016, xiaolei_low-dimensional_2019, sandip_comparative_2018}, level2]
        [Contextual Similarity \cite{a_learning_2013, hamid_deep_2015, nemanja_hierarchical_2015}, level2]
    ]
    [From Sparse \\ to Dense \\ Representations, level1,
        [One-Hot Encoding \cite{yong_intrinsic_2018, a_learning_2013, rachith_inexplicable_2023}, level2]
        [Word Embeddings \cite{a_learning_2013, xiaolei_low-dimensional_2019, matthias_context_2018}, level2]
    ]
    [Contextualized \\ Word Embeddings, level1,
        [ELMo \cite{matthew_deep_2018}, level2]
        [BERT and Variants \cite{devlin2018bert, liu2019roberta, lan2019albert}, level2]
        [{Other Contextualized Embeddings:} \\ {GPT, XLNet, XLM} \cite{radford2018improving, yang2019xlnet, conneau2019cross}, level2]
    ]
    [Subword-Level \\ Embeddings and \\ Generalization, level1,
        [Handling Rare and Unseen \\ Words \cite{hamed_embedding-based_2016, debasis_word_2015, jinman_generalizing_2018, a_learning_2013, mourad_word_2022}, level2]
        [Cross-Lingual Embeddings \cite{peng_investigating_2016, trevor_cross-lingual_2017, fangxiaoyu_language-agnostic_2020, will_bilingual_2013}, level2]
    ]
    [Personalized \\ Word Embeddings, level1,
        [Modeling Linguistic Variation \cite{charles_exploring_2020, yossi_analysis_2017}, level2]
        [Applications \cite{charles_exploring_2020, hamid_deep_2015, doddapaneni_user_2024}, level2]
    ]
]
\end{forest}
    \caption{Taxonomy of Word Embeddings}
    \label{fig:taxonomy-word}
\end{figure*}

\section{Word Embeddings and Language Models}
\subsection{Foundational Concepts}
\subsubsection{\textbf{Distributional Hypothesis}}
The distributional hypothesis, a cornerstone of numerous word embedding techniques, posits that words appearing in similar contexts tend to have similar meanings \cite{lebret_word_2016}. This hypothesis allows for representing words as vectors in a continuous space, where semantic similarity is reflected by vector proximity \cite{xiaolei_low-dimensional_2019}. This shift from symbolic to distributed representations has revolutionized NLP, enabling advancements in tasks like information retrieval, machine translation, and sentiment analysis \cite{sandip_comparative_2018}. The distributional hypothesis has limitations, however. It often struggles to capture the nuances of word meaning in different contexts, particularly for polysemous words, leading to a need for context-dependent representations \cite{tammie_introduction_2021}. Additionally, while effective at capturing broad semantic and syntactic relations, traditional word embeddings based on the distributional hypothesis can be computationally expensive and struggle with issues like the curse of dimensionality, out-of-vocabulary words, and overfitting to specific domains \cite{matthias_context_2018}. Moreover, focusing primarily on word properties rather than morphology can lead to inconsistent performance across different evaluation metrics \cite{yong_intrinsic_2018}. This has motivated research into alternative word embedding models that incorporate subword information, particularly for morphologically rich languages \cite{peng_investigating_2016}.

\subsubsection{\textbf{Contextual Similarity}} Context plays a critical role in disambiguating word meanings and improving the performance of language models. Different definitions of context offer unique approaches to capture semantic relationships. Local context, often defined as a sliding window of neighboring words, is utilized by models like Word2Vec to learn word embeddings through the prediction of nearby words \cite{a_learning_2013}. Although efficient, this approach can be limited in its ability to capture long-range dependencies. Sentence-level context, which considers the entire sentence, allows models like LSTM-RNNs to integrate more comprehensive information, resulting in more nuanced sentence embeddings \cite{hamid_deep_2015}. Document-level context further expands the scope, encompassing the entire document, which proves advantageous for tasks such as document classification and in representing temporal relationships between documents in sequential data streams \cite{nemanja_hierarchical_2015}.

Contextualized word embeddings, generated by models such as ELMo and BERT, incorporate contextual information directly into word representations and enable the same word to have distinct embeddings based on its context \cite{shao-yen_multimodal_2019}. This contextualization is crucial for addressing polysemy and enhancing performance in tasks like word sense disambiguation (WSD) \cite{matthew_deep_2018}. For instance, contextual string embeddings, which represent words as character sequences and integrate surrounding text, have proven especially effective for WSD, notably in named entity recognition tasks \cite{a_contextual_2018}. Additionally, the type and extent of context used can substantially influence language model performance. For example, while adding socio-situational information offers slight improvements to character-based models, it may adversely affect models using word embeddings, underscoring the need to carefully assess the interplay between context and embedding type \cite{linda_effect_2019}. This has driven research into refined context selection strategies and data augmentation techniques aimed at further enhancing language model capabilities \cite{matthias_scope_2018}.

\subsection{From Sparse Representations to Dense Representations}
\subsubsection{One-Hot Encoding}
One-hot encoding represents words as sparse, high-dimensional vectors. Each word in the vocabulary is assigned a unique vector in which only one element is set to one, while all others remain zero. Despite its simplicity, this representation has several critical limitations. The dimensionality of these vectors scales linearly with the vocabulary size, leading to extremely high dimensions for large vocabularies \cite{yong_intrinsic_2018}. This results in significant sparsity, with most elements being zero, making computations inefficient and resource-intensive \cite{a_learning_2013}. Additionally, one-hot encoding lacks the ability to capture semantic relationships between words. Each word is represented independently, with no intrinsic way to reflect similarities or relationships between words. This is a substantial limitation because capturing semantic and syntactic relationships is crucial for many natural language processing (NLP) tasks \cite{yang_word_2018}. For instance, as discussed in \cite{rachith_inexplicable_2023}, in an n-gram language model, predicting the next word in a sequence relies on word co-occurrence. If certain words never co-occur, the model cannot infer any relationship between them. Thus, one-hot encoding fails to capture the contextual similarity needed for effective language modeling and other NLP tasks, such as semantic search, knowledge base question answering, and machine translation \cite{hamed_embedding-based_2016, japa_question_2020}.

\subsubsection{Word Embeddings}
Word embeddings address the limitations of one-hot encoding by representing words as dense, low-dimensional vectors learned from large corpora \cite{a_learning_2013}. These vectors capture semantic and syntactic relationships, mapping words into a continuous vector space where similar words are positioned closer together \cite{xiaolei_low-dimensional_2019}. This arrangement is consistent with the distributional hypothesis, which states that words occurring in similar contexts tend to have similar meanings \cite{marco_dont_2014}. This contextual information is valuable for a range of NLP tasks, including word similarity measurement \cite{yong_intrinsic_2018}, analogy solving \cite{andrew_modeling_2015}, document retrieval \cite{hamid_deep_2015}, and machine translation \cite{will_bilingual_2013}.

The continuous vector space representation also allows algebraic operations on word embeddings. This property is particularly useful in tasks such as analogy solving, where relationships between words can be expressed as vector operations \cite{sanjeev_latent_2015}. By embedding words into a latent space, these models capture the underlying semantic and syntactic structure of language \cite{matthias_context_2018}. This latent representation is especially advantageous for downstream tasks like question answering, named entity recognition, and neural machine translation, as it enables models to generalize beyond specific word occurrences and recognize broader semantic relationships. Unlike one-hot encoding, word embeddings mitigate the problems of high dimensionality, sparsity, and the lack of semantic representation \cite{yang_word_2018, tammie_introduction_2021}. The dense vectors allow more efficient computations and enable the model to learn complex relationships based on the distributional properties of large text corpora \cite{mourad_word_2022}.

Word embeddings extend beyond individual words to represent phrases and even entire documents, as shown in \cite{shaohua_embedding_2016} and \cite{mohammad_embeddings_2021}. This versatility makes word embeddings a powerful and flexible tool for a wide range of NLP applications.

\subsection{Contextualized Word Embeddings}
\subsubsection{\textbf{ELMo}}
ELMo \cite{matthew_deep_2018} employs bidirectional LSTMs trained on a language modeling objective to generate contextualized word representations. Unlike static word embeddings (e.g., Word2Vec, GloVe) that assign a single vector per word regardless of context, ELMo produces dynamic embeddings that vary based on the word's surrounding text. This contextual sensitivity allows ELMo to capture nuanced meanings and disambiguate polysemous words. The architecture consists of two LSTM layers stacked on top of each other, each processing the input sequence in both forward and backward directions. The internal states of these LSTMs at each time step are combined, often through a weighted average, to form the contextualized word representation. Different layers of the bidirectional LSTM capture different aspects of linguistic information. The lower layers tend to encode syntactic information, while higher layers focus on semantic aspects. Alternative weighting schemes for combining these layers can be learned during downstream task training, optimizing the contribution of each layer for specific tasks.

\subsubsection{BERT and Variants}
BERT \cite{devlin2018bert} and other transformer-based bidirectional encoders use a Transformer encoder architecture and are pre-trained on two objectives: Masked Language Modeling (MLM) and Next Sentence Prediction (NSP). MLM randomly masks tokens in the input sequence and trains the model to predict the masked words based on the surrounding context. NSP trains the model to determine whether two given sentences are consecutive in the original text. These objectives enable BERT to capture bidirectional context effectively, addressing limitations of previous models that relied on unidirectional or shallow bidirectional representations. Fine-tuning BERT for specific downstream tasks involves adding task-specific layers on top of the pre-trained encoder and training the entire model on labeled data for the target task.

RoBERTa \cite{liu2019roberta} modifies the BERT pre-training procedure by removing NSP, training with larger batch sizes and more data, and dynamically changing the masking pattern during training. ALBERT \cite{lan2019albert} introduces parameter reduction techniques, such as factorized embedding parameterization and cross-layer parameter sharing, to reduce the model size and improve training efficiency while maintaining performance. These variants address some of the computational challenges associated with BERT and offer improved performance on various downstream tasks. BERT and its variants handle polysemy by generating distinct embeddings for a word based on its context. The MLM objective helps capture the meaning of rare words by leveraging the surrounding context. However, long sequences can still pose challenges due to the quadratic complexity of the self-attention mechanism in the Transformer architecture. The tying of word vectors and word classifiers as a loss framework for language modeling \cite{hakan_tying_2016} has also been explored as a way to improve language modeling performance.

\subsubsection{Other Contextualized Embeddings: GPT, XLNet, and XLM}
GPT \cite{radford2018improving} utilizes a Transformer decoder architecture and is trained using a language modeling objective, predicting the next word in a sequence. This autoregressive approach captures dependencies in one direction, limiting its ability to capture full bidirectional context. XLNet \cite{yang2019xlnet} addresses this limitation by using a permutation language modeling objective, which considers all possible orderings of the input sequence during training, allowing it to capture bidirectional context while maintaining the autoregressive formulation. XLM \cite{conneau2019cross} extends BERT to support cross-lingual training, utilizing a translation language modeling objective to learn representations that capture relationships between words in different languages. The input embeddings of these models typically represent word-level information, while the output embeddings encode contextualized representations. Tying input and output embeddings \cite{hakan_tying_2016} can be used to reduce the number of parameters and potentially improve performance. Using the output embedding to improve language models has been explored in various contexts, often by incorporating it into downstream tasks or as a way to refine the language model itself.

\subsection{Subword-Level Word Embeddings and Generalization}

\subsubsection{Subword Information for Handling Rare and Unseen Words}
Standard word embedding techniques often struggle with rare and unseen words, resulting in out-of-vocabulary (OOV) issues that hinder generalization to new vocabulary. Subword-level information, such as character n-grams and morphemes, offers a solution to this limitation by representing words as compositions of subword units. With this approach, subword embedding models can create meaningful representations even for words not encountered during training.

\cite{hamed_embedding-based_2016} examined word embeddings to expand queries with semantically related terms, enhancing the accuracy and robustness of query language models in handling vocabulary mismatches. Building on this, \cite{debasis_word_2015} proposed a generalized language model utilizing word embeddings to address vocabulary mismatch in information retrieval. This model conceptualizes query term observation as a two-step process: generating an intermediate term from the document or collection and transforming it into the observed query term, thus effectively capturing term relationships and mitigating vocabulary mismatches.

In a similar vein, \cite{jinman_generalizing_2018} introduced a subword-level vector model that treats words as bags of character n-grams, allowing it to generate effective embeddings for rare or unseen words. This model is computationally efficient, easy to train, and demonstrates state-of-the-art performance on word similarity and morphosyntactic tasks across multiple languages. Complementing this approach, \cite{a_learning_2013} presented a simplified method for training word embeddings on rare or unseen words, using noise-contrastive estimation for log-bilinear models. This method is faster and more efficient than prior models, producing better results in terms of embedding quality and generalization. Additionally, \cite{mourad_word_2022} offers an extensive review of word embedding models, including subword-level approaches, and their applications across various NLP tasks.

\subsubsection{Cross-Lingual Word Embeddings and Low-Resource Languages}
Subword information is instrumental in advancing cross-lingual word embeddings, particularly for low-resource languages with limited training data. By capturing morphological similarities across languages, subword-level models can learn shared representations for morphologically related words, even when they exhibit different surface forms. \cite{peng_investigating_2016} explored universal and language-specific properties within word embeddings, revealing that word form features are particularly beneficial for inflectional languages. Similarly, \cite{trevor_cross-lingual_2017} examined cross-lingual word embeddings derived from bilingual lexicons to enhance language models for low-resource languages. Applying this approach to Yongning Na, they highlighted challenges and potential solutions for low-resource settings.

Furthermore, \cite{fangxiaoyu_language-agnostic_2020} developed a language-agnostic BERT model, LaBSE, which supports over 100 languages and sets a new benchmark for cross-lingual tasks. LaBSE leverages both multilingual and monolingual data, using techniques like MLM and TLM to learn effective cross-lingual sentence embeddings. Finally, \cite{will_bilingual_2013} introduced bilingual word embeddings based on a large, unlabeled corpus and machine translation word alignments, showing notable improvements in machine translation by capturing semantic equivalence across languages.

\subsection{Personalized Word Embeddings}
\subsubsection{Modeling Individual Linguistic Variation}

Personalized word embeddings aim to capture individual differences in word usage and linguistic preferences, advancing beyond standard generic representations. These embeddings can enhance language model performance in tasks customized to specific users. \cite{charles_exploring_2020} investigated the value of personalized word embeddings in language modeling, finding that combining generic and personalized embeddings led to a 4.7\% relative reduction in perplexity, thus improving model performance. They also observed that words associated with specific psycholinguistic categories showed greater variation across users, indicating that personalized models are particularly advantageous for predicting such words. Additionally, \cite{yossi_analysis_2017} examined properties encoded within sentence embeddings, providing insights into the influence of factors such as word frequency and positional distance on a model’s encoding of content and word order. Their findings on individual linguistic variation at the sentence level offer valuable guidance for developing more user-specific word embeddings and NLP systems.

\subsubsection{Applications of Personalized Embeddings}

Personalized word embeddings have proven beneficial across various NLP tasks. \cite{charles_exploring_2020} demonstrated the potential of personalized embeddings in authorship attribution, showing that they can effectively capture unique writing styles. \cite{hamid_deep_2015} introduced LSTM-RNN models for sentence embedding trained on user click-through data, underscoring the application of personalized embeddings in user-specific document retrieval systems that provide more relevant search results and tailored recommendations. Personalized embeddings can also leverage user history data: \cite{doddapaneni_user_2024} proposed a User Embedding Module (UEM) that compresses user histories into embeddings to serve as prompts in language models. This approach yielded improvements in personalized language tasks, suggesting that personalized embeddings effectively capture user preferences and interests for customized recommendations and user profiling.

\noindent \textbf{Fig \ref{fig:taxonomy-word}} presents a taxonomy of word embeddings discussed in this section. It categorizes embeddings based on their underlying techniques and applications, ranging from traditional one-hot encoding and dense word embeddings to advanced contextualized and personalized embeddings.

\section{Sentence and Document Embeddings}
\subsection{Sentence Embeddings}
Sentence embeddings capture the semantic meaning of entire sentences as fixed-length vectors, which are fundamental for many natural language processing (NLP) tasks, including semantic search, text classification, and question answering. Various approaches have been developed to generate sentence embeddings, ranging from simple averaging methods to more complex neural architectures.

\subsubsection{Simple Averaging and Pooling of Word Embeddings}
A straightforward approach to generating sentence embeddings involves combining the word embeddings of individual words within a sentence through simple averaging or pooling techniques. These methods take advantage of pre-trained word embeddings, which are widely available across different languages.

\begin{itemize}[leftmargin=*] \item \textbf{Methods}: Common methods include averaging the word embeddings, or using element-wise maximum (max-pooling), minimum (min-pooling), or sum operations. These approaches have been explored in various contexts, such as in a comparative study of word embedding models for hate speech detection \cite{minni_comparison_2021}. \item \textbf{Strengths}: The main advantage of these methods is their simplicity and computational efficiency. They are easy to implement and require minimal computational resources, making them ideal for large-scale applications. For example, a study comparing bag-of-words approaches with sentence encoders found that simple averaging can be unexpectedly effective for certain tasks \cite{c_evaluation_2018}. \item \textbf{Limitations}: However, these methods have notable limitations. Averaging or pooling word embeddings typically results in a loss of word order information, which limits their effectiveness in tasks where word order is significant. Additionally, they can be sensitive to outliers, as an uncommon word may disproportionately influence the resulting sentence embedding. To mitigate these issues, researchers have proposed more advanced techniques, such as using linear combinations of word embeddings to capture higher-level linguistic structures \cite{lyndon_surprising_2019}. \end{itemize}

\subsubsection{Recurrent Neural Network (RNN) based Approaches}
Recurrent Neural Networks (RNNs) provide a more advanced approach to sentence embedding by sequentially processing words and capturing dependencies across them. RNN variants, such as Long Short-Term Memory (LSTM) cells and Gated Recurrent Units (GRUs), are especially effective for generating contextual embeddings.

\begin{itemize}[leftmargin=*] \item \textbf{Architectures}: LSTM-RNNs and GRUs are commonly used for sentence embedding as they effectively capture long-range dependencies within sentences. These architectures surpass simple averaging methods in tasks that require deeper contextual understanding. For instance, a study on deep sentence embedding with LSTM networks demonstrated their capacity to detect keywords and allocate topics autonomously \cite{hamid_deep_2015}.
\item \textbf{Training}: RNN-based models can be trained using various approaches, including weakly supervised methods. One example involves training LSTM-RNNs with user click-through data from web search engines, which helps models learn effective representations for document retrieval \cite{hamid_deep_2015}.

\item \textbf{Applications}: RNN-generated sentence embeddings have been applied to numerous NLP tasks, such as document retrieval and semantic similarity measurement. For instance, using LSTM-RNN embeddings for document retrieval yielded notable performance improvements over traditional methods \cite{hamid_deep_2015}.
\end{itemize}

\subsubsection{Transformer-based Sentence Encoders}

\paragraph{Architecture}
Sentence-BERT (SBERT) has become a leading architecture for generating sentence embeddings, as highlighted by \cite{a_knowledge_2020}. SBERT leverages pre-trained transformer models like BERT, fine-tuning them specifically for sentence-level tasks to generate efficient and effective sentence embeddings. Another notable model, SBERT-WK \cite{bin_sbert-wk_2020}, refines BERT-based word models to produce high-quality sentence representations. This model focuses on analyzing word representation evolution across BERT layers, applying geometric analysis to enhance sentence embeddings.

\paragraph{Fine-tuning Strategies}
Fine-tuning strategies are essential for adapting pre-trained language models to sentence embedding tasks. A common method is supervised fine-tuning on Natural Language Inference (NLI) and Semantic Textual Similarity (STS) datasets, as discussed in \cite{fangxiaoyu_language-agnostic_2020}. This approach uses labeled sentence pairs with semantic relationship annotations (e.g., entailment, contradiction, similarity) to refine model performance. Knowledge distillation also proves effective; \cite{nils_making_2020} demonstrates how a student model can be trained to mimic a teacher model, effectively transferring insights from a high-resource language to multiple target languages.

\paragraph{Performance}
Transformer-based sentence encoders achieve state-of-the-art results in various Semantic Textual Similarity (STS) tasks \cite{j_bilingual_2019}. These models excel at capturing semantic relationships, making them highly effective for tasks requiring understanding of sentence meaning and similarity. Specifically, SBERT-WK demonstrates superior performance on STS benchmarks, further validating its robustness in sentence embedding applications \cite{bin_sbert-wk_2020}.

\paragraph{Word Representation Evolution Across Layers}
Analyzing the evolution of word representations across transformer layers provides insights into how these models capture and process contextual information. Studies such as \cite{r_lmfingerprints_2022} introduce methods to quantify and visualize contextualization across layers, often analyzing changes in word embeddings and examining the unique information captured by each layer. Additionally, \cite{steven_analysing_2020} investigates word representations at both input and output stages, offering deeper insights into layer-wise contextual processing.

\paragraph{Incorporating Backward Dependencies for Enhanced Semantic Similarity (BeLLM)}
The BeLLM model \cite{xianming_bellm_2023} addresses the limitations of unidirectional attention in autoregressive large language models by introducing backward dependencies. By transforming specific attention layers to be bidirectional, BeLLM enhances semantic similarity tasks by considering both preceding and following context, thereby providing a more comprehensive contextual understanding.

\subsubsection{Multilingual and Cross-lingual Sentence Embeddings}

\paragraph{Knowledge Distillation}
Knowledge distillation is an effective technique for aligning vector spaces across languages, facilitating cross-lingual understanding \cite{nils_making_2020}. By training a student model to replicate the vector space of a monolingual teacher model, semantic knowledge from a high-resource language can be transferred to other languages, including those with limited resources.

\paragraph{Translation Ranking and Bi-text Mining}
Dual encoder architectures and negative sampling are core techniques in translation ranking and bi-text mining tasks, as explored by \cite{fangxiaoyu_language-agnostic_2020}. In these architectures, source and target sentences are encoded separately, with training focused on ranking true translation pairs higher than negative samples, thereby improving translation accuracy and relevance.

\paragraph{Language-agnostic BERT Sentence Embedding (LaBSE)}
The Language-agnostic BERT Sentence Embedding (LaBSE) model \cite{fangxiaoyu_language-agnostic_2020} has demonstrated high performance in multilingual applications. LaBSE utilizes a combination of pre-training and dual-encoder fine-tuning to optimize translation ranking and produce language-agnostic sentence embeddings that excel across languages.

\paragraph{Addressing Language Bias in Multilingual Embeddings}
Addressing language bias is essential to ensure fair and effective multilingual embeddings. \cite{nils_making_2020} notes that some models may exhibit biases, favoring specific language combinations over others, which can impair performance in multilingual tasks. Techniques such as knowledge distillation are valuable for mitigating this bias and achieving more balanced cross-lingual representations.

\paragraph{Challenges in Low-Resource Languages}
Handling low-resource languages remains challenging in multilingual NLP. \cite{trevor_cross-lingual_2017} highlights that bilingual lexicons can provide an effective solution by enabling cross-lingual word embedding generation when text data is scarce. This approach enhances language model performance in low-resource environments, helping bridge the gap in multilingual understanding.

\subsection{Document Embeddings}

Document embeddings extend the concept of word embeddings to represent entire documents as fixed-length vectors. These representations capture the semantic meaning of larger text units, supporting various downstream tasks such as document classification, retrieval, and clustering. This section reviews key techniques for creating document embeddings, with a focus on methods for combining word embeddings and exploring generative topic embedding models.

\subsubsection{Methods for Combining Word Embeddings}

Creating document embeddings often involves aggregating word embeddings from individual words in the document. This section discusses common aggregation methods, including averaging, pooling, and other compositional techniques.

\paragraph{Averaging and Pooling: Methods, Strengths, and Limitations}
A straightforward approach to creating document embeddings is to average the word embeddings of all words in the document. Although simple, this method can be highly effective and sometimes outperforms more complex models, particularly when word order is not critical \cite{lyndon_surprising_2019}. Averaging captures semantic relationships by leveraging word co-occurrence patterns \cite{lyndon_surprising_2019}. However, this method overlooks word order and may miss nuanced information in complex sentences \cite{lyndon_surprising_2019}.

Pooling methods, such as max-pooling or min-pooling, offer an alternative by selecting the most salient features from word embeddings. While these methods can highlight important features, they may still lose significant contextual information. More sophisticated techniques, like using recurrent neural networks (RNNs) with Long Short-Term Memory (LSTM) cells, process words sequentially, accumulating richer information as they traverse the document \cite{hamid_deep_2015}. This method has shown effectiveness in tasks like document retrieval, as demonstrated in \cite{hamid_deep_2015}.

\paragraph{Compositional Methods for Aggregating Word Embeddings}
Beyond simple averaging and pooling, compositional methods aim to capture interactions between words within a document. \cite{mo_factor-based_2014} explores compositional embedding models that create representations for substructures within annotated sentences, addressing the challenge of representing a diverse range of sentences by building sentence embeddings from component word embeddings. Additionally, \cite{lebret_word_2016} presents an approach that learns embeddings for phrases by combining word embeddings, providing richer phrase-level representations for documents.

\subsubsection{Generative Topic Embedding Models}

Generative topic embedding models combine the strengths of word embeddings and topic modeling to learn latent document representations. These models overcome some limitations of traditional topic models by capturing both local and global semantic information.

\paragraph{Combining Word Embeddings with Topic Modeling (LDA)}
Integrating word embeddings into topic models, such as Latent Dirichlet Allocation (LDA), enhances the quality of learned topics. \cite{shaohua_embedding_2016} proposes a generative topic embedding model that merges local word collocation patterns with global document-level topic patterns through variational inference. This approach yields topic embeddings and topic mixing proportions for each document, resulting in low-dimensional, continuous representations \cite{shaohua_embedding_2016}.

\paragraph{Correlated Topic Models for Semantic Relatedness}
Correlated topic models aim to capture relationships between topics. \cite{guangxu_correlated_2017} introduces the Correlated Gaussian Topic Model (CGTM), which uses word embeddings to represent topics as multivariate Gaussian distributions. This structure captures semantic relatedness and correlations between words, improving topic coherence over traditional topic models \cite{guangxu_correlated_2017}.

\paragraph{Embedded Topic Model (ETM) for Enhanced Topic Quality}
The Embedded Topic Model (ETM), described in \cite{adji_topic_2019}, combines word embeddings with traditional topic models to improve topic quality and predictive performance. Unlike LDA, ETM uses word and topic embeddings to model word probabilities within a topic, effectively handling large vocabularies and rare words. This model has demonstrated superior performance over LDA in settings with extensive vocabularies \cite{adji_topic_2019}.

\subsubsection{Hierarchical Neural Language Models for Documents}

Hierarchical neural language models provide a robust approach for learning distributed representations of both documents and words in data streams. These models use multi-layer architectures, where each layer captures distinct levels of information.

\paragraph{Multi-layer Architectures for Document and Word Representations}
Hierarchical neural models often use layered structures to capture different aspects of context. \cite{nemanja_hierarchical_2015} presents a two-layer model: the upper layer models the temporal context of document sequences, assuming that temporally close documents are likely to be related, while the lower layer captures word sequence context within each document. This setup enables the model to learn joint representations for both documents and words within a shared feature space, which simplifies tasks like document tagging through nearest-neighbor search. This hierarchical structure can be extended with additional layers, such as a user layer, to incorporate user preferences for personalized recommendations.

\paragraph{Temporal Context in Document and Word Sequences}
A key advantage of hierarchical neural models is their ability to incorporate temporal context. \cite{nemanja_hierarchical_2015} shows how these models learn representations that account for temporal co-occurrences of documents within a stream. By optimizing the joint log-likelihood of document and word sequences, the model balances context at both levels, moving beyond traditional bag-of-words approaches. This temporal awareness supports a more dynamic understanding of language, resulting in improved performance in tasks like movie classification (MovieLens dataset) and click-through prediction (Yahoo News dataset).

\paragraph{Applications: Document Retrieval, Recommendation, and Tagging}
The joint representations of documents and words learned by hierarchical neural language models enable a range of applications. For document retrieval, the similarity between a query and document can be efficiently computed by comparing their vector representations in the shared embedding space. In recommendation systems, the model leverages user preferences, represented in an additional user layer, to generate personalized suggestions. For automatic tagging, the shared embedding space facilitates nearest-neighbor searches to assign relevant tags based on document vectors. \cite{nemanja_hierarchical_2015} demonstrates the model’s superior performance on tasks like movie classification, underscoring its versatility in document retrieval and recommendation. The ability to capture both document- and word-level context makes these models highly applicable for text mining.

\begin{figure*}
    \centering
    
\tikzset{
    basic/.style  = {draw, align=left, font=\sffamily, rectangle},
    root/.style   = {basic, rounded corners=2pt, thin, fill=red!30, text width=2.2cm, anchor=west},
    level1/.style = {basic, thin, rounded corners=2pt, fill=red!20, text width=3.7cm},
    level2/.style = {basic, thin, rounded corners=2pt, fill=red!10, text width=8.2cm},
    leaf/.style   = {basic, thin, fill=white!80!red, text width=5cm},
    edge from parent/.style={draw=black, edge from parent fork right},
    level distance=2cm,
}

\begin{forest}
for tree={
    grow=east,
    scale=0.8, 
    growth parent anchor=west,
    parent anchor=east,
    child anchor=west,
    l sep=6mm, 
    s sep=2mm, 
    edge path={
        \noexpand\path[\forestoption{edge},->, >={latex}] 
        (!u.parent anchor) -- +(5pt,0pt) |- (.child anchor) 
        \forestoption{edge label};
    },
    align=left, 
}
[Sentence \\ and \\ Document \\ Embeddings, root,
    [Cross-Lingual and \\ Multi-Lingual \\ Embeddings, level1,
        [Cross-Lingual Word Embeddings \cite{will_bilingual_2013, trevor_cross-lingual_2017}, level2]
        [Challenges and Evaluation \cite{takashi_unsupervised_2018, sebastian_survey_2017}, level2]
        [Language-Agnostic Sentence Embeddings \cite{nils_making_2020, fangxiaoyu_language-agnostic_2020}, level2]
    ]
    [Document Embeddings, level1,
        [Methods for Combining Word Embeddings \\ \cite{lyndon_surprising_2019, hamid_deep_2015, mo_factor-based_2014, lebret_word_2016}, level2]
        [Generative Topic Embedding Models \cite{shaohua_embedding_2016, guangxu_correlated_2017, adji_topic_2019}, level2]
        [Hierarchical Neural Language Models \cite{nemanja_hierarchical_2015}, level2]
    ]
    [Sentence Embeddings, level1,
        [Simple Averaging and Pooling \cite{minni_comparison_2021, c_evaluation_2018, lyndon_surprising_2019}, level2]
        [Recurrent Neural Networks \cite{hamid_deep_2015}, level2]
        [Transformer-based Sentence Encoders \\ \cite{a_knowledge_2020, bin_sbert-wk_2020, fangxiaoyu_language-agnostic_2020, nils_making_2020, r_lmfingerprints_2022, steven_analysing_2020, xianming_bellm_2023}, level2]
    ]
]
\end{forest}
    \caption{Taxonomy of Sentence and Document Embeddings}
    \label{fig:taxonomy-sentence-document}
\end{figure*}
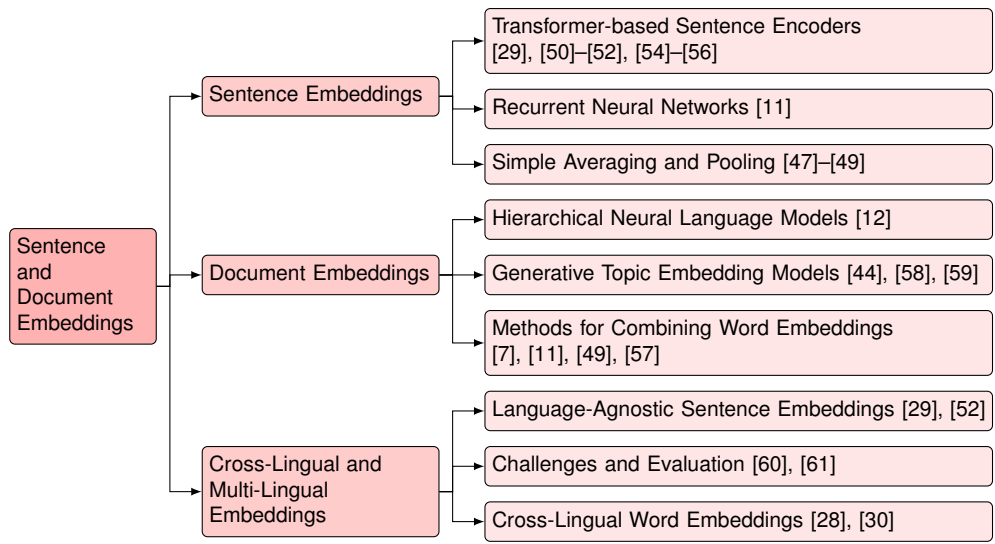

\subsection{Cross-Lingual and Multi-Lingual Embeddings}

\subsubsection{Cross-Lingual Word Embeddings: Methods and Applications}

Cross-lingual word embeddings map words from different languages into a shared vector space, facilitating cross-lingual knowledge transfer. This approach is especially useful for low-resource languages, which often lack sufficient monolingual training data. Supervised methods typically use bilingual lexicons or word alignments from parallel corpora. For example, \cite{will_bilingual_2013} introduces bilingual word embeddings learned from a large, unlabeled corpus, where machine translation word alignments are used to enforce translational equivalence. These embeddings, when integrated into a phrase-based machine translation system, show significant improvements in translation quality. Another supervised approach, explored in \cite{trevor_cross-lingual_2017}, leverages bilingual lexicons to enhance language models for low-resource languages, applying this technique to the language Yongning Na.

In contrast, unsupervised methods develop cross-lingual embeddings without explicit parallel data. \cite{takashi_unsupervised_2018} proposes a multilingual neural language model that shares parameters across languages, jointly learning word embeddings in a shared space without parallel data or pre-training. This method proves effective even with limited monolingual data or domain differences across languages. For a comprehensive overview of contextual and cross-lingual embeddings, \cite{qi_survey_2020} offers valuable insights.

\subsubsection{Challenges and Evaluation of Cross-Lingual Embeddings}

Developing cross-lingual embeddings entails significant challenges, especially in low-resource settings and when data domains differ across languages. Limited training data can impair unsupervised methods, as noted in \cite{takashi_unsupervised_2018}, where application to the language Yongning Na highlights resource scarcity challenges. Domain differences across languages also impact embedding quality; while \cite{takashi_unsupervised_2018} demonstrates that their model can handle different-domain corpora, achieving high-quality embeddings remains challenging.

Evaluating cross-lingual embeddings typically involves both intrinsic and extrinsic tasks. Intrinsic evaluations assess embeddings through word similarity tasks, while extrinsic evaluations examine embedding performance in downstream applications. For instance, \cite{will_bilingual_2013} uses the NIST08 Chinese-English machine translation task for extrinsic evaluation. For a thorough discussion on the challenges and evaluation of cross-lingual embeddings, \cite{sebastian_survey_2017} provides extensive analysis.

\subsubsection{Language-Agnostic Sentence Embeddings: Techniques, Evaluation, and Addressing Bias}

Language-agnostic sentence embeddings represent sentences from different languages in a unified vector space, enabling cross-lingual semantic similarity and retrieval tasks. Techniques for generating these embeddings include knowledge distillation and fine-tuning of pre-trained multilingual models. \cite{nils_making_2020} proposes extending monolingual sentence embeddings to new languages via knowledge distillation, where a student model replicates the vector space of a monolingual teacher model. This approach shows strong results for low-resource languages and effectively minimizes language bias. Additionally, \cite{fangxiaoyu_language-agnostic_2020} introduces LaBSE, a language-agnostic BERT sentence embedding model that achieves state-of-the-art performance on bi-text retrieval tasks.

Evaluation of language-agnostic sentence embeddings often involves tasks like semantic textual similarity (STS) and bi-text mining. Addressing language bias, where models may preferentially align certain language pairs, is a key challenge. \cite{nils_making_2020} notes that their approach, as well as mUSE, avoid language bias issues seen in models like LASER and LaBSE.
\newline 
\noindent \textbf{Fig \ref{fig:taxonomy-sentence-document}} illustrates the taxonomy of sentence and document embeddings discussed in this section. The taxonomy categorizes embedding techniques into three main areas: Sentence Embeddings, Document Embeddings, and Cross-Lingual and Multi-Lingual Embeddings. Each category is further subdivided into specific methodologies.

\section{Grounding Language Models in Other Modalities}

\subsection{Grounding Language to Vision}

\subsubsection{Connecting Language and Perception}
With the advancing capabilities of language models, researchers are increasingly exploring their applications in multimodal tasks, particularly those involving vision. Grounding language models in visual perception bridges the gap between linguistic representations and the physical world. This approach is inspired by how humans learn language—by associating words and concepts with perceptual experiences. 

By grounding language models in vision, we can create more robust and comprehensive representations that capture not only statistical relationships between words but also their connections to tangible objects and scenes. This alignment of language with perception is essential for developing AI systems capable of understanding and interacting with the world in a more human-like manner. Such multimodal grounding enables applications like image captioning, visual question answering, and robotic manipulation, where an AI system’s ability to interpret both language and visual context is crucial.

\subsubsection{Methods for Visual Grounding}

\paragraph{Joint Embedding Spaces: Learning Shared Representations for Images and Text}
A common approach to visual grounding involves creating joint embedding spaces for images and text. In these spaces, visual and textual data are mapped into a shared vector representation, enabling direct comparison and interaction between the two modalities. \cite{ryan_unifying_2014} introduces an encoder-decoder pipeline to achieve this. In this architecture, an encoder maps both images and sentences into a shared latent space, and a decoder generates descriptions from this space. Another approach utilizes multimodal contrastive learning, as described in \cite{yizhen_explainable_2021}, where a two-stream model for visual and language inputs aligns representations by maximizing agreement between paired image-text data while minimizing agreement for unrelated pairs.

\paragraph{Attention Mechanisms for Cross-Modal Retrieval and Alignment}
Attention mechanisms are highly effective for aligning visual and textual information in cross-modal retrieval tasks. By employing attention, models can identify the most relevant visual features for a textual query or vice versa. \cite{yizhen_explainable_2021} employs a cross-modal attention module that retrieves specific visual objects based on language queries. This mechanism allows for fine-grained alignment, enabling the model to focus on essential parts of the input specific to a task.

\paragraph{Similarity Learning and Grounding Object, Verb, and Attribute Representations}
Similarity learning plays an essential role in grounding representations of objects, verbs, and attributes. \cite{ghaffari_grounding_2023} uses similarity learning within an embodied simulation environment, enabling comparison of objects based on their properties and behaviors. By learning a similarity metric through interactions within this simulation, the model effectively grounds word vectors to object representations, facilitating tasks like zero-shot learning, where novel objects can be recognized based on descriptions alone. Notably, \cite{ghaffari_grounding_2023} also found that grounding object representations enhances the grounding of verbs and attributes, aligning with findings in analogical reasoning.

\subsubsection{Applications of Visually Grounded Language Models}

Visually grounded language models, which learn joint representations of images and text, have shown considerable promise in multimodal applications. These models align visual and textual information, allowing them to excel in tasks that require understanding the relationship between images and their corresponding descriptions.

\paragraph{Image Captioning and Retrieval}
Image captioning generates textual descriptions for images, while image retrieval involves finding images that match a given textual query. \cite{ryan_unifying_2014} introduced an encoder-decoder pipeline that creates a joint embedding space for images and text, combined with a "structure-content" neural language model. This model achieved state-of-the-art results in both image captioning and retrieval tasks on the Flickr8K and Flickr30K datasets.

\paragraph{Visual Question Answering (VQA)}
Visual Question Answering (VQA) requires answering questions based on visual information within images. \cite{shao-yen_multimodal_2019} explored multimodal embeddings by integrating language models with acoustic information and paralinguistic cues to enhance emotion recognition—a crucial aspect for VQA applications. Additionally, \cite{jaeyong_deep_2015} developed a deep neural network that learns a shared embedding space for point-cloud, natural language, and manipulation trajectory data, enabling robots to manipulate novel objects based on instructions and visual experience. \cite{yue_language_2021} introduced the Visual Semantic Embedding Probe (VSEP) to evaluate contextualized word embeddings for zero-shot object recognition in complex visual scenes. Further, \cite{singh2023emb} utilized pre-trained language models to create interpretable and efficient predictors for tasks, including visually grounded applications.

\paragraph{Referring Expression Comprehension and Generation}
Referring expression comprehension identifies an object in an image based on a textual description, while referring expression generation produces a unique textual description for a specified object within an image. \cite{peng_kosmos-2_2023} introduced Kosmos-2, a multimodal large language model capable of grounding text to visual information, such as drawing bounding boxes around objects. Trained on a large-scale dataset of grounded image-text pairs, Kosmos-2 demonstrated enhanced performance on tasks involving referring expression comprehension and generation.

\paragraph{Multimodal Image Search}
Multimodal image search uses a blend of textual and visual queries to locate images. \cite{yizhen_explainable_2021} proposed a two-stream model that aligns visual and language representations through cross-modal contrastive learning, enabling multimodal image searches using queries composed of images, text, or combinations of both.

\subsubsection{Impacts and Challenges}

Visually grounded language models have demonstrated benefits across a range of linguistic tasks. \cite{shahmohammadi_language_2022} investigated the impact of visual grounding on both concrete and abstract words, finding that grounding can enhance representations for both types of words. This study also highlighted the advantages of visual grounding for contextualized embeddings, particularly when trained on smaller corpora, where the added visual context helps compensate for limited textual data. \cite{matthias_context_2018} examined the role of contextual information in language modeling, showing that context is essential for addressing semantic challenges in tasks like word sense disambiguation.

Despite these advancements, visually grounded language models face notable challenges. \cite{yue_language_2021} identified the lack of mutual exclusivity bias as a limitation in current visual semantic embedding models, where models may fail to distinguish between overlapping or closely related objects within scenes. Addressing such biases and enabling language models to reason more effectively about novel visual scenes are key areas for future research, with the goal of enhancing the models’ robustness and generalization across diverse visual-linguistic contexts.

\subsection{Multimodal Embeddings for Robotics}

\subsubsection{Motivation: Challenges in Representing and Integrating Diverse Sensor Modalities}

Robots operating in unstructured, real-world environments face significant challenges in processing and integrating data from diverse sensor modalities. These include visual data from cameras and depth sensors, natural language instructions from human operators, haptic feedback, and motion trajectories from manipulators. Integrating these varied data streams is essential for robots to build a comprehensive understanding of their environment and interact effectively. However, as highlighted by \cite{jaeyong_deep_2015}, manually designing features that bridge these modalities is extremely difficult due to differences in data formats and inherent complexities. For instance, visual data may be represented as images, point clouds, or voxel grids, each with unique dimensional structures, whereas language instructions are symbolic sequences that must be translated into a format suitable for robotic control. Similarly, manipulation trajectories can be represented as time series data of joint angles, end-effector positions, or force/torque measurements. This heterogeneity presents a substantial barrier to direct comparison or fusion of information across modalities.

Developing techniques that can automatically learn joint representations to capture relationships between these diverse modalities is crucial for advancing robot autonomy and intelligence. Additionally, as noted by \cite{matthias_context_2018}, context is essential for interpreting both linguistic and sensorimotor information. While embeddings are effective in capturing syntactic and some semantic relationships, they require contextualization to handle challenges such as word sense disambiguation and the integration of real-world knowledge, particularly in robotics applications.

\subsubsection{Methods for Multimodal Embeddings in Robotics}

\paragraph{Deep Neural Networks for Shared Embedding Spaces: Architectures and Training Strategies}
Deep neural networks are a powerful tool for learning shared embedding spaces that unify different modalities, mapping them into a common vector space where semantically related concepts are closer together. Various architectures are used for this purpose. For example, \cite{yizhen_explainable_2021} discusses a two-stream architecture where separate networks process individual modalities, such as vision and language, before merging outputs into a joint representational space. Alternatively, a single network can concurrently process all modalities, directly learning a shared embedding space.

Training strategies for multimodal embeddings typically involve a two-stage process: an initial pre-training phase on large unimodal datasets, such as image-caption pairs or text corpora, to learn general features within each modality, followed by fine-tuning on smaller multimodal datasets to align representations across modalities. This approach proved effective in \cite{shao-yen_multimodal_2019}, where acoustic information was integrated into contextualized lexical embeddings to enhance emotion recognition through paralinguistic cues.

\paragraph{Loss Functions for Multimodal Similarity Learning: Margin-based Approaches}
Effective multimodal embedding training relies on carefully designed loss functions. Margin-based approaches, which aim to separate embeddings of dissimilar concepts while grouping similar ones, are commonly used. For instance, triplet loss, as discussed by \cite{shahmohammadi_language_2022}, teaches the model to position an anchor instance closer to a positive (semantically similar) instance than to a negative (semantically dissimilar) instance by a predefined margin. Similarly, contrastive loss, employed in \cite{jaeyong_deep_2015}, works by distinguishing between similar and dissimilar pairs, pushing dissimilar embeddings apart while pulling similar ones closer. These loss functions encourage a semantically structured embedding space that effectively captures relationships between modalities.

\subsubsection{Combining Point-Cloud Data, Natural Language Instructions, and Manipulation Trajectories}
The integration of point-cloud data, natural language instructions, and manipulation trajectories is critical for enabling robots to understand and execute complex tasks in real-world environments. Point clouds, obtained from depth sensors, provide 3D geometric information about the environment. Natural language instructions specify goals and actions, while manipulation trajectories, derived from demonstrations or learned experience, offer actionable paths to achieve these goals.

Embedding these diverse modalities into a shared space, as explored in \cite{jaeyong_deep_2015}, enables the establishment of meaningful connections across representations. For instance, robots can ground language instructions, such as "grasp the blue object," by linking the instruction to the corresponding point cloud of a blue object and its associated manipulation trajectory. This multimodal grounding enables generalization to novel objects and instructions, fostering more flexible and adaptive robotic behaviors.

\subsubsection{Applications of Multimodal Embeddings in Robotics}

\paragraph{Robotic Manipulation of Novel Objects}
Multimodal embeddings enable robots to interact with and manipulate unfamiliar objects based on previous experiences. By learning a shared embedding space that includes object representations (e.g., point clouds), language instructions, and manipulation trajectories, as demonstrated by \cite{jaeyong_deep_2015}, robots can generalize their knowledge to novel objects. When encountering a new object alongside a language instruction, the robot can map the object's representation into the shared embedding space, retrieving the most relevant manipulation trajectory aligned with the instruction. This capability allows robots to adapt to new situations and perform tasks without requiring explicit programming for every new object-action combination.

\paragraph{Task Planning and Execution Based on Natural Language Instructions}
Comprehending and executing complex tasks specified via natural language is a critical objective in robotics. Multimodal embeddings facilitate this by aligning language commands with corresponding action sequences. This enables robots to understand and perform multi-step tasks expressed in natural language, such as "navigate to the table, pick up the book, and place it on the shelf." With a multimodal embedding model, the robot can translate these instructions into a sequence of actions, including navigation, object recognition, grasping, and placement. This approach moves beyond simple, keyword-based command interpretation, fostering more flexible and sophisticated human-robot interactions. Future research, as suggested by \cite{japa_question_2020}, could enhance this capability by fine-tuning language models with attention mechanisms to improve the accuracy and robustness of mapping natural language instructions to specific actions.

\paragraph{Evaluation and Benchmarks: Real-World Robot Experiments and Simulated Environments}
Evaluating multimodal embeddings in robotics requires a mix of real-world experiments and simulations. Real-world experiments provide a realistic measure of a model’s ability to generalize to new scenarios and handle the complexities of unstructured environments. However, these tests can be resource-intensive and time-consuming. Simulated environments offer a controlled and efficient platform to evaluate model performance across various tasks and conditions. For instance, \cite{jaeyong_deep_2015} demonstrates how simulations can pre-train models before deploying them on physical robots. Common evaluation metrics include task success rate, execution time, and path efficiency.

\paragraph{Open Challenges and Future Directions: Scalability and Real-Time Performance}
The field of multimodal embeddings for robotics presents ongoing challenges and exciting research directions. One primary challenge is scalability to high-dimensional sensor data. As robots incorporate advanced sensors like high-resolution cameras and lidar, the dimensionality of data increases significantly. Efficiently processing and embedding this high-dimensional data is essential for real-world applications. Another critical challenge is achieving real-time performance; real-time sensor data processing and action generation are vital for effective interaction in dynamic environments. Continued research is needed to create more efficient embedding models and architectures that support real-time robotic control. Exploring alternative embedding techniques, such as those discussed by \cite{a_learning_2013}, could yield scalable, efficient solutions for handling high-dimensional data in real-time applications.

\subsection{Modeling Brain Activity with Language Model Embeddings}

\subsubsection{Motivation: Connecting Artificial and Biological Language Processing}

The impressive capabilities of large language models (LLMs) in natural language processing tasks have prompted research into their internal representations and whether these mirror the way the human brain processes language. \cite{yossi_analysis_2017} discusses the shift from word-level to sentence-level representations in language models, which reflects advances in understanding complex language representations. Exploring the connection between artificial and biological language processing is essential for developing AI that is more human-like and for gaining insights into the cognitive foundations of language comprehension. \cite{richard_low-dimensional_2021} investigates how the structure of language representations correlates with human brain responses, highlighting the potential for identifying shared computational principles. Additionally, \cite{matthias_context_2018} emphasizes the role of embeddings and contextual information, which could enhance our understanding of both artificial and biological language processing systems. This line of research not only enriches cognitive neuroscience but also informs the development of more robust and interpretable language models.

\subsubsection{Methods for Modeling Brain Activity}

\paragraph{Neuroimaging Techniques (e.g., fMRI, EEG, ECoG) for Capturing Brain Responses to Language}
Several neuroimaging techniques allow researchers to study brain activity during language processing. Functional magnetic resonance imaging (fMRI) measures brain activity through blood flow changes, providing insight into the brain regions involved in various language tasks. \cite{shailee_incorporating_2018} used fMRI to analyze the impact of contextual information on brain responses. Electroencephalography (EEG) captures electrical activity from scalp electrodes, offering high temporal resolution for tracking neural response timing to language stimuli. \cite{andrea_family_2023} applied EEG to examine brain responses to familiar entities. Electrocorticography (ECoG), which involves placing electrodes on the brain’s surface, provides even higher spatial resolution than EEG. \cite{ariel_brain_2022} utilized ECoG to study the geometric similarity between brain embeddings and contextual language model embeddings. These techniques capture neural responses to linguistic stimuli, ranging from single words to complex narratives.

\paragraph{Extracting Brain Embeddings from Neural Recordings}
Brain embeddings are continuous vector representations derived from neural activity in response to linguistic stimuli. These embeddings encapsulate patterns of brain responses across specific time windows and regions of interest (ROIs). For example, \cite{ariel_brain_2022} extracted brain embeddings by averaging neural activity in the inferior frontal gyrus (IFG) during a listening task. This enables direct comparison with word embeddings derived from language models, facilitating a deeper understanding of how the brain processes language.

\paragraph{Comparing Brain Embeddings to Language Model Embeddings: Geometric Alignment and Zero-Shot Mapping}
Techniques such as geometric alignment and zero-shot mapping are used to compare brain embeddings with language model embeddings. Geometric alignment assesses the structural similarity between embedding spaces, while zero-shot mapping involves predicting brain activity patterns for new words based on their language model representations, or vice versa. \cite{richard_low-dimensional_2021} adapted a transfer learning approach to reveal a low-dimensional structure in language model representations that aligns with brain responses. \cite{ariel_brain_2022} found that brain embeddings and DLM embeddings share similar geometry, enabling zero-shot predictions of brain activity for novel words, suggesting a potential alignment between human and artificial language representation.

\subsubsection{Relating Language Model Layers to Temporal Dynamics of Brain Activity}

Research has explored the relationship between the layered architecture of deep language models (DLMs) and the temporal dynamics of brain activity. \cite{goldstein_correspondence_2022} found a correspondence between sequential transformations in DLM layers and the temporal sequence of neural activity in brain regions associated with language. This suggests that different DLM layers may reflect different stages of language processing in the brain. However, the study also identifies unique neural processes, such as recurrent information accumulation and responses to unpredictable words, which are not fully captured by current DLMs.

\subsubsection{Encoding Personal Memories and Experiences with Language Models: Subject-Specific Semantic Representations}

While language models are typically trained on large general corpora, individual experiences shape human language understanding. \cite{andrea_family_2023} demonstrated that language models could represent subject-specific semantic knowledge of familiar people and places when trained on personalized data. This capability opens up possibilities for personalized language models that capture both general and individual semantic knowledge, with applications in personalized recommendations and user-specific content generation.

\subsubsection{Open Challenges and Future Directions: Dimensionality of Brain Embedding Spaces, Neural Processing of Unpredictable Words, and Linking Brain Embeddings to Cognitive Processes}

Despite recent advancements, challenges remain in modeling brain activity with language model embeddings. Understanding the dimensionality of brain embedding spaces and how they relate to language model dimensions is essential. Another key challenge is modeling neural processes for unpredictable words, which current models do not fully capture. Additionally, linking brain embeddings to cognitive processes and behavioral measures could enhance our understanding of the cognitive mechanisms underlying language. \cite{yossi_analysis_2017} underscores the need to understand properties encoded in sentence representations, which could inform future research in this area. Addressing these challenges may lead to more cognitively plausible language models and deeper insights into the neural basis of language.

\section{Advanced Topics and Research Gaps}

\subsection{Embedding Model Compression: Techniques for Reducing Model Size and Memory Footprint}

\subsubsection{Need for Compression: Computational and Storage Challenges of Large Embedding Matrices}
The impressive performance of large language models (LLMs) in NLP applications often comes with substantial computational and storage requirements, particularly due to the embedding layers that represent vocabulary items as dense vectors. As vocabularies expand to include hundreds of thousands of tokens and embedding dimensions increase to capture nuanced semantic relationships, the memory footprint can become prohibitive, especially in resource-constrained environments like mobile devices. This scalability challenge is further compounded by the increasing demand for multilingual and multimodal models, which necessitate even larger embedding matrices to represent diverse linguistic and perceptual information \cite{shao-yen_multimodal_2019, fangxiaoyu_language-agnostic_2020}. Efficient compression techniques are essential for effectively deploying these powerful models \cite{dan_fighting_2018}. Additionally, the computational cost associated with large embedding matrices can impede training and inference speed, impacting real-time applications \cite{a_learning_2013, mourad_word_2022}.

\subsubsection{Compression Techniques}

\paragraph{Knowledge Distillation: Training Smaller Student Models to Mimic Larger Teacher Models}
Knowledge distillation trains a smaller "student" model to replicate the behavior of a larger "teacher" model, effectively transferring knowledge from a complex model to a more efficient one. In cross-lingual embeddings, this technique helps transfer the rich semantic space of a high-resource language model to a smaller model that supports multiple languages, as demonstrated in \cite{nils_making_2020}. This approach is particularly valuable for low-resource languages by leveraging knowledge from data-rich languages \cite{trevor_cross-lingual_2017}. Knowledge distillation can also be applied for vocabulary reduction, enabling a smaller embedding matrix to capture essential semantic information with a condensed vocabulary \cite{sanqiang_extreme_2019}.

\paragraph{Weight Tying: Sharing Parameters Between Input and Output Embeddings}
Weight tying reduces model size by sharing parameters between different layers, commonly between input and output embedding matrices in language models \cite{hakan_tying_2016, ofir_using_2016}. This approach not only decreases the memory footprint but can also enhance performance by promoting consistency between input and output representations. \cite{kristina_how_2018} further proposes a modification that decouples the hidden state from word prediction, achieving comparable or better results with fewer parameters.

\paragraph{Quantization: Reducing Embedding Vector Precision}
Quantization reduces the precision of embedding vectors (e.g., using 8-bit integers instead of 32-bit floats), significantly lowering storage requirements and accelerating computations, though potentially at the cost of precision loss. Quantization techniques are particularly effective for memory-constrained devices \cite{tao2022compression}.

\paragraph{Pruning: Removing Less Important Embedding Dimensions or Vectors}
Pruning techniques reduce model size by removing redundant or less important dimensions or vectors from the embedding matrix. By carefully selecting pruning criteria, this method can maintain performance while making the model more efficient, though excessive pruning can degrade performance \cite{wang2019structured}.

\paragraph{Subspace Embedding Methods: Representing Embeddings in Lower-Dimensional Subspaces}
Subspace embedding methods map high-dimensional embeddings into lower-dimensional subspaces. For instance, \cite{amit_lightweight_2023} uses a Cartesian product of subspaces to construct compact embeddings, while \cite{xu_tensorgpt_2023} applies Tensor-Train Decomposition (TTD) to compress embedding layers, achieving significant size reduction with minimal or no performance loss. Additionally, \cite{kartik_diverse_2014} projects word history vectors onto multiple diverse low-dimensional subspaces.

\subsubsection{Evaluating Compression Methods: Balancing Compression Rate and Task Performance}
Evaluating embedding compression techniques involves balancing the compression rate against performance on downstream tasks. High compression rates improve efficiency but can adversely impact model accuracy. \cite{sanqiang_extreme_2019} demonstrated significant compression with minimal performance drop, while \cite{amit_lightweight_2023} achieved high compression with only a slight decrease in accuracy. Selecting an appropriate compression method requires careful consideration of the target application and tolerance for performance degradation.

\subsection{Interpretability and Explainability of Embeddings}

\subsubsection{The Black Box Nature of Embedding Spaces}
Embedding models have revolutionized natural language processing by capturing semantic relationships between words, phrases, and sentences. However, the internal mechanisms of these models remain largely opaque, creating challenges for interpretability. Although embeddings perform well in various tasks, understanding why they work and what specific information they encode remains difficult. This "black box" nature limits our ability to fully trust and effectively debug these models. Traditional word embeddings, for instance, can struggle with nuanced word-level relationships \cite{steven_analysing_2020}. Transformer-based models, despite excelling at complex NLP tasks, present additional interpretability challenges due to their depth and complexity, making it difficult to understand the properties encoded in sentence representations and potential risks of information leakage \cite{yossi_analysis_2017}. As models grow more sophisticated, developing techniques to interpret their embeddings becomes increasingly critical.

\subsubsection{Methods for Understanding Embeddings}

\paragraph{Visualization Techniques}
Visualizing embedding spaces provides insights into relationships between words, sentences, and concepts. Dimensionality reduction techniques like Principal Component Analysis (PCA), t-distributed Stochastic Neighbor Embedding (t-SNE), and Uniform Manifold Approximation and Projection (UMAP) are often used to project high-dimensional embeddings into two or three dimensions for visualization. For example, \cite{dima_comparative_2018} visualized word embeddings to understand how Arabic word semantics are represented. Similarly, \cite{rishabh_comparative_2021} applied dimensionality reduction to visualize the semantic space of text in the sports domain. However, visualizations come with limitations: projecting high-dimensional data into lower dimensions involves information loss, and the choice of technique can affect the insights gained.

\paragraph{Probing Tasks}
Probing tasks are designed to evaluate specific linguistic properties within embeddings. These tasks involve training a simple classifier (e.g., linear classifier) to predict linguistic attributes (e.g., part-of-speech, tense) from embeddings, with classifier performance indicating the extent to which embeddings capture these attributes. For instance, \cite{alessio_contextual_2020} used probing tasks to compare the linguistic knowledge in BERT and Word2vec embeddings. Additionally, \cite{matthew_dissecting_2018} analyzed how word representation varies across layers in neural language models, finding that deeper layers capture complex linguistic features like semantics and morphology. \cite{steven_analysing_2020} further explored how input and output embeddings in neural models encode different semantic knowledge.

\paragraph{Concept Activation Vectors (CAVs)}
Concept Activation Vectors (CAVs) offer insights into the contribution of specific concepts to a model's predictions. A CAV represents a direction in the embedding space corresponding to a particular concept. By projecting embeddings onto this direction, we can measure the association of embeddings with that concept. \cite{tennenholtz_demystifying_2023} briefly references CAVs but does not provide specifics on their implementation.

\paragraph{Embedding Inversion Attacks}
Embedding inversion attacks aim to reconstruct the original text from its embedding, highlighting privacy risks and potential information leakage. \cite{li_sentence_2023} proposed a generative inversion attack that could reconstruct input sequences from sentence embeddings, exposing the possibility of recovering sensitive information from anonymized representations.

\subsubsection{Using LLMs to Interpret Embeddings}
Large language models (LLMs) present a novel approach to interpreting embeddings by generating explanations and narratives for embedding vectors. As demonstrated in \cite{tennenholtz_demystifying_2023}, LLMs can translate abstract embeddings into understandable descriptions, enhancing interpretability and expanding applications for embeddings.

\subsubsection{Applications of Explainable Embeddings}
Explainable embeddings are essential for developing more trustworthy AI systems and improving model debugging. By understanding the encoded information and its impact on predictions, we can identify and mitigate biases \cite{shiva_evaluating_2023}. Explainable embeddings also aid in debugging by clarifying why a model made specific predictions, allowing identification and correction of errors or biases in the model’s training data or architecture. While interpretability remains a challenge, ongoing research in methods like visualization, probing tasks, CAVs, embedding inversion attacks, and leveraging LLMs for interpretation, are paving the way for more transparent and reliable language models.

\subsection{Encoding Numerical Information in Language Models}

\subsubsection{Challenges of Representing Numbers in Text}
Representing numbers poses unique challenges for language models due to their inherent quantitative properties, which differ from typical lexical tokens. Unlike words, numbers have magnitudes and relationships that are essential for understanding quantitative context. For example, the difference between "1" and "2" is a quantitative one, not just a lexical difference, and the relationship between "1" and "10" is distinct from that between "9" and "10," despite similar lexical distances. Standard word embedding methods often treat numbers as regular tokens, failing to capture these crucial quantitative and relational aspects. This limitation becomes particularly problematic for tasks requiring numerical reasoning or quantitative information extraction, where the model must understand magnitudes and relationships between numbers \cite{t_language_2021}.

\subsubsection{Embedding Numerical Information}

\paragraph{Default Embedding Methods: Treating Numbers as Regular Tokens}
In most language models, numbers are treated as standard tokens and are assigned vector representations similar to words. This approach does not inherently encode the quantitative properties of numbers, often resulting in poor performance on tasks involving numerical magnitudes or relationships \cite{andrew_modeling_2015}. The limitations of this method are most apparent in mathematical reasoning and quantitative information extraction tasks, where an understanding of numbers is crucial.

\paragraph{Exponential Embeddings: Encoding Numerical Magnitude}
Exponential embeddings seek to address these limitations by encoding numerical magnitudes in the embedding space, often using a logarithmic scale. This approach allows the embedding distance to approximate the ratio between numbers, capturing some aspects of numerical magnitude. However, while exponential embeddings offer an improvement over default methods, they still struggle to fully represent numerical relationships \cite{linda_effect_2019}.

\paragraph{Floating-Point Embeddings: Leveraging Computer System Representations}
Floating-point embeddings take inspiration from the computer's floating-point representation, encoding numbers as a combination of a mantissa and an exponent. This enables models to capture both the magnitude and precision of numerical values, improving their comprehension of numerical information and enhancing performance on numerical reasoning tasks \cite{xiaoxiao_floating-point_2024}.

\subsubsection{Specialized Architectures for Numerical Reasoning in LLMs}
To bolster numerical reasoning capabilities in large language models (LLMs), researchers are developing specialized architectures that integrate external knowledge sources or modules for handling numerical and mathematical operations. Such architectures provide promising advancements for tasks requiring numerical reasoning, quantitative information extraction, and other forms of numerical understanding \cite{qiang_bilstm_2021}.

\subsubsection{Evaluation of Numerical Reasoning Capabilities: Datasets and Metrics}
Evaluating the numerical reasoning abilities of language models requires specialized datasets and metrics. Datasets like Numeracy-600K and Wiki-Concert contain textual data with numerical information, serving as benchmarks to assess model performance on numerical reasoning tasks \cite{xiaoxiao_floating-point_2024}. Key metrics—such as accuracy, precision, recall, and F1-score—help quantify a model’s capability to correctly extract and process numerical information. These datasets and metrics are crucial for tracking advancements in developing language models with strong numerical reasoning capabilities.

\subsection{Efficient and Scalable Training of Embedding Models}

\subsubsection{Computational Cost of Training Large Embedding Matrices}
The increasing scale of data used in natural language processing (NLP) has led to a proportional increase in embedding matrix sizes, creating significant computational challenges. This is particularly evident in pre-trained language models like BERT and GPT, which often contain billions of parameters. The memory footprint of these models is substantial, posing difficulties for deployment on resource-constrained devices and impacting both training and inference times. For instance, \cite{yong_intrinsic_2018} emphasizes the need for efficient language representations and improved evaluation methods, while \cite{sanqiang_extreme_2019} focuses on compressing large models like BERT to reduce computational demands.

\subsubsection{Efficient Training Techniques}
Several techniques aim to reduce the computational cost of training large embedding matrices while preserving the quality of the learned embeddings \cite{lebret_word_2016}. These methods focus on optimizing the training process to achieve faster convergence and lower resource consumption.

\paragraph{Noise-Contrastive Estimation: Approximating the Loss Function for Faster Training}
Noise-contrastive estimation (NCE) approximates the loss function in training word embeddings, significantly accelerating the process. \cite{a_learning_2013} demonstrates how NCE enables high-quality word embedding training with reduced data and computational needs by simplifying loss computation, especially beneficial for large vocabularies and datasets.

\paragraph{Subsampling Techniques: Reducing the Number of Training Examples for Faster Convergence}
Subsampling reduces the number of training examples used in the embedding learning process, facilitating faster convergence and lower computational costs. \cite{stephan_training_2016} presents an efficient subsampling approach for neural word embeddings trained with hierarchical softmax, focusing on the most informative examples for quicker and potentially more accurate learning.

\paragraph{Optimized Batching Strategies: Improving Training Throughput with Large Batch Sizes}
Optimized batching strategies enhance training throughput by employing larger batch sizes, which is particularly useful for large language models. \cite{chen_bge_2024} discusses using large batches to improve embedding discriminativeness. Larger batch sizes lead to more stable parameter updates and faster convergence, although careful tuning is required to manage memory constraints and convergence stability.

\paragraph{Parallel and Distributed Training Methods: Scaling Up on Multiple GPUs or CPUs}
Parallel and distributed training techniques distribute computational workload across multiple GPUs or CPUs, drastically reducing training time for large models and datasets. \cite{andrew_modeling_2015} describes parallel training methods that enable scaling beyond previous model sizes, facilitating training for very large neural networks.

\subsubsection{Hardware Acceleration for Embedding Training and Inference}
Specialized hardware such as GPUs and TPUs offers significant acceleration for both training and inference of embedding models. These devices’ parallel processing capabilities are ideally suited for the matrix operations integral to embedding computations, expediting model experimentation and deployment in NLP applications \cite{raha2023efficient}.

\subsection{Addressing Bias and Ethical Concerns in Embeddings}

\subsubsection{Sources of Bias in Embedding Models}
Biases in language models pose ethical challenges and can perpetuate societal inequalities. These biases arise from various sources, including the training data and the model architecture. Training data often reflects societal biases, as noted by \cite{yong_intrinsic_2018}. If the corpus overrepresents specific demographics or viewpoints, the resulting embeddings may inherit and amplify these biases. Model architecture also contributes to bias; design choices, such as the objective function or the incorporation of context, may unintentionally favor certain types of information, leading to skewed representations. For instance, models that prioritize word co-occurrence patterns may inadvertently capture and reinforce biases present in the data \cite{yossi_analysis_2017}.

\subsubsection{Measuring and Quantifying Bias}
Accurate measurement and quantification of bias are essential for understanding its nature and impact. Intrinsic and extrinsic evaluation metrics are commonly used. Intrinsic evaluations directly assess embeddings for bias, often through similarity measures or analogy tasks. For instance, \cite{felix_not_2014} used word similarity tasks to compare translation-based and monolingual embeddings, revealing differences in how they capture ontological information. Extrinsic evaluations measure bias by examining its impact on downstream NLP tasks. Statistical tests, like the Word Embedding Association Test (WEAT), help quantify biased associations, while \cite{bin_evaluating_2019} used correlation analysis to explore the relationship between intrinsic and extrinsic evaluations. Analyzing bias's differential impact on social groups is also essential; \cite{shiva_evaluating_2023} investigated intersectional biases, examining how gender identity, sexual orientation, and social class intersect to influence biased representations.

\subsubsection{Mitigating Bias}
Bias mitigation in language models is an active area of research. Data augmentation can balance training datasets, minimizing the effects of skewed representations \cite{matthias_context_2018}. Debiasing techniques aim to neutralize biased associations within embedding spaces; for instance, \cite{t_language_2021} proposed a new metric for evaluating vector transformations. Additionally, adversarial training helps models become robust to biased inputs by making them insensitive to specific protected attributes, enhancing fairness in model predictions.

\subsubsection{Ethical Implications of Biased Embeddings}
Biased embeddings raise ethical concerns surrounding fairness, transparency, and accountability in NLP systems. Embeddings that perpetuate stereotypes or discriminate against certain groups can lead to harmful societal consequences. For example, \cite{minni_comparison_2021} examined word embeddings for hate speech detection, emphasizing the importance of capturing context for accurate identification. Transparency is essential to understand the origins of biases and their impact on model decisions, while accountability ensures that those developing and deploying NLP systems take responsibility for mitigating bias and addressing potential harms.

\subsection{Adaptive Language Modeling and Transfer Learning with Embeddings}

\subsubsection{Transfer Learning: Using Pre-trained Embeddings for Downstream Tasks}
Pre-trained word embeddings have become a cornerstone of NLP, providing rich semantic information derived from extensive unlabeled text. This transfer learning approach has proven effective across diverse NLP applications. For instance, \cite{hamid_deep_2015} demonstrates that LSTM-RNN sentence embeddings enhance web document retrieval performance, while \cite{g_integrating_2015} shows how neural word embeddings improve translation model effectiveness. Beyond retrieval tasks, \cite{yossi_analysis_2017} analyzes properties encoded in sentence embeddings, offering insights into their capture of content, order, and length. Pre-trained embeddings also benefit other areas: \cite{debasis_word_2015} employs word embeddings to tackle vocabulary mismatch in information retrieval, yielding notable improvements, and \cite{japa_question_2020} highlights the effectiveness of BERT embeddings in knowledge base question answering.

\subsubsection{Domain Adaptation: Fine-tuning Embeddings on Domain-Specific Data}
While generic pre-trained embeddings provide a strong baseline, fine-tuning on domain-specific data can further enhance their utility, particularly for specialized vocabularies and semantic nuances. \cite{daniel_language_2021} demonstrates that embeddings trained on engineering texts outperform general-purpose models in engineering-related tasks. In the clinical domain, \cite{yuqi_enhancing_2019} finds that contextual embeddings pre-trained on clinical data achieve state-of-the-art performance in clinical concept extraction. Similarly, \cite{linda_effect_2019} explores the impact of domain-specific context on RNN language models, emphasizing the benefits of domain adaptation.

\subsubsection{Cross-Lingual Transfer Learning: Leveraging Embeddings for Low-Resource Languages}
Cross-lingual embeddings enable knowledge transfer from resource-rich to resource-poor languages by projecting words into a shared embedding space. \cite{trevor_cross-lingual_2017} explores cross-lingual embeddings from bilingual lexicons to improve low-resource language models, while \cite{will_bilingual_2013} introduces bilingual embeddings learned from unlabeled corpora and machine translation alignments, enhancing translation performance. Even with limited data, \cite{felix_embedding_2014} shows that embeddings from neural machine translation models outperform monolingual embeddings by capturing conceptual and lexical-syntactic similarities. \cite{takashi_unsupervised_2018} further proposes an unsupervised method for cross-lingual embeddings using multilingual language models, achieving high quality with limited monolingual data and across varying domains.

\subsubsection{Adaptive Language Modeling: Conditioning Language Models on Different Contexts}
Adaptive language modeling allows models to condition on specific contexts, such as style or sentiment, for more controlled text generation. \cite{nils_alternative_2019} explores weighting schemes for ELMo embeddings to improve downstream performance in varied contexts. \cite{chi_word_2023} introduces LM-Switch, a lightweight approach for adapting language models to specific conditions by manipulating word embeddings, enabling nuanced language generation.

\subsubsection{Zero-Shot Learning with Embeddings: Applying Language Models to New Tasks without Task-Specific Training Data}
Zero-shot learning with embeddings enables language models to tackle new tasks without task-specific training data. \cite{yue_language_2021} introduces the Visual Semantic Embedding Probe (VSEP) for evaluating contextualized word embeddings in zero-shot visual semantic understanding, while \cite{nancy_rethinking_2020} argues for evaluating language model embeddings on their direct linguistic reasoning capabilities, showing potential for zero-shot common-sense reasoning tasks.

\begin{figure*}
    \centering
    
\tikzset{
    basic/.style  = {draw, align=left, font=\sffamily, rectangle},
    root/.style   = {basic, rounded corners=2pt, thin, fill=green!30, text width=3.2cm, anchor=west},
    level1/.style = {basic, thin, rounded corners=2pt, fill=green!20, text width=4cm},
    level2/.style = {basic, thin, rounded corners=2pt, fill=green!10, text width=6cm},
    edge from parent/.style={draw=black, edge from parent fork right},
    level distance=2cm,
}

\begin{forest}
for tree={
    grow=east,
    scale=0.7, 
    growth parent anchor=west,
    parent anchor=east,
    child anchor=west,
    l sep=6mm, 
    s sep=1mm, 
    edge path={
        \noexpand\path[\forestoption{edge},->, >={latex}] 
        (!u.parent anchor) -- +(5pt,0pt) |- (.child anchor) 
        \forestoption{edge label};
    },
    align=left, 
}
[Advanced Topics \\ and Research Gaps, root,
    [Embedding Model \\ Compression, level1,
        [Compression Techniques, level2,
            [Knowledge Distillation \cite{nils_making_2020, trevor_cross-lingual_2017, sanqiang_extreme_2019}, level2]
            [Weight Tying \cite{hakan_tying_2016, ofir_using_2016, kristina_how_2018}, level2]
            [Quantization \cite{tao2022compression}, level2]
            [Pruning \cite{wang2019structured}, level2]
            [Subspace Embedding Methods \\ \cite{amit_lightweight_2023, xu_tensorgpt_2023, kartik_diverse_2014}, level2]
        ]
        [Evaluating Compression Methods \\ \cite{sanqiang_extreme_2019, amit_lightweight_2023}, level2]
    ]
    [Interpretability and \\ Explainability \\ of Embeddings, level1,
        [The Black Box Nature \cite{steven_analysing_2020, yossi_analysis_2017}, level2]
        [Methods for Understanding \\ Embeddings, level2,
            [Visualization Techniques \cite{dima_comparative_2018, rishabh_comparative_2021}, level2]
            [Probing Tasks \cite{alessio_contextual_2020, matthew_dissecting_2018, steven_analysing_2020}, level2]
            [Concept Activation Vectors \\ (CAVs) \cite{tennenholtz_demystifying_2023}, level2]
            [Embedding Inversion Attacks \cite{li_sentence_2023}, level2]
        ]
        [Using LLMs to Interpret Embeddings, level2]
        [Explainable Embeddings \cite{shiva_evaluating_2023}, level2]
    ]
    [Encoding \\ Numerical Information \\ in Language Models, level1,
        [Challenges of \\ Representing Numbers \cite{t_language_2021}, level2]
        [Embedding Numerical Information, level2,
            [Default Embedding Methods \cite{andrew_modeling_2015}, level2]
            [Exponential Embeddings \cite{linda_effect_2019}, level2]
            [Floating-Point Embeddings \cite{xiaoxiao_floating-point_2024}, level2]
        ]
        [Specialized Architectures for \\ Numerical Reasoning \cite{qiang_bilstm_2021}, level2]
        [Evaluation of Numerical \\ Reasoning Capabilities \cite{xiaoxiao_floating-point_2024}, level2]
    ]
    [Efficient and Scalable \\ Training of \\ Embedding Models, level1,
        [Computational \\ Cost of Training \cite{yong_intrinsic_2018, sanqiang_extreme_2019}, level2]
        [Efficient Training \\ Techniques \cite{a_learning_2013, stephan_training_2016, chen_bge_2024, andrew_modeling_2015}, level2,
            [Noise-Contrastive Estimation \cite{a_learning_2013}, level2]
            [Subsampling Techniques \cite{stephan_training_2016}, level2]
            [Optimized Batching Strategies \cite{chen_bge_2024}, level2]
            [Parallel and Distributed \\ Training Methods \cite{andrew_modeling_2015}, level2]
        ]
        [Hardware Acceleration \\ for Embedding Training \\ and Inference \cite{raha2023efficient}, level2]
    ]
    [Addressing \\ Bias and Ethical \\ Concerns in Embeddings, level1,
        [Sources of Bias in \\ Embedding Models \cite{yong_intrinsic_2018, yossi_analysis_2017}, level2]
        [Measuring and Quantifying Bias \\ \cite{felix_not_2014, bin_evaluating_2019, shiva_evaluating_2023}, level2]
        [Mitigating Bias \cite{matthias_context_2018, t_language_2021}, level2]
        [Ethical Implications \\ of Biased Embeddings \cite{minni_comparison_2021}, level2]
    ]
    [Adaptive \\ Language Modeling and \\ Transfer Learning \\ with Embeddings, level1,
        [Transfer Learning \\ \cite{hamid_deep_2015, g_integrating_2015, yossi_analysis_2017, debasis_word_2015, japa_question_2020}, level2]
        [Domain Adaptation \cite{daniel_language_2021, yuqi_enhancing_2019, linda_effect_2019}, level2]
        [Cross-Lingual Transfer Learning \\ \cite{trevor_cross-lingual_2017, will_bilingual_2013, felix_embedding_2014, takashi_unsupervised_2018}, level2]
        [Adaptive Language Modeling \\ \cite{nils_alternative_2019, chi_word_2023}, level2]
        [Zero-Shot Learning with \\ Embeddings \cite{yue_language_2021, nancy_rethinking_2020}, level2]
    ]
    [The Role of Embeddings \\ in Emerging Areas, level1,
        [Embodied AI \cite{jaeyong_deep_2015, driess_palm-e_2023, peng_kosmos-2_2023}, level2]
        [Cognitive Science \\ \cite{ariel_brain_2022, niu2024large, ariel_alignment_2024, andrea_family_2023, goldstein_correspondence_2022, yossi_analysis_2017}, level2]
        [Reasoning and \\ Commonsense Knowledge \\ \cite{mirza_language_2022, rajat_complementary_2020, mourad_word_2022, nancy_rethinking_2020, sanjeev_latent_2015}, level2]
    ]
]
\end{forest}
    \caption{Taxonomy of Advanced Topics and Research Gaps in Word Embeddings}
    \label{fig:taxonomy_advanced}
\end{figure*}
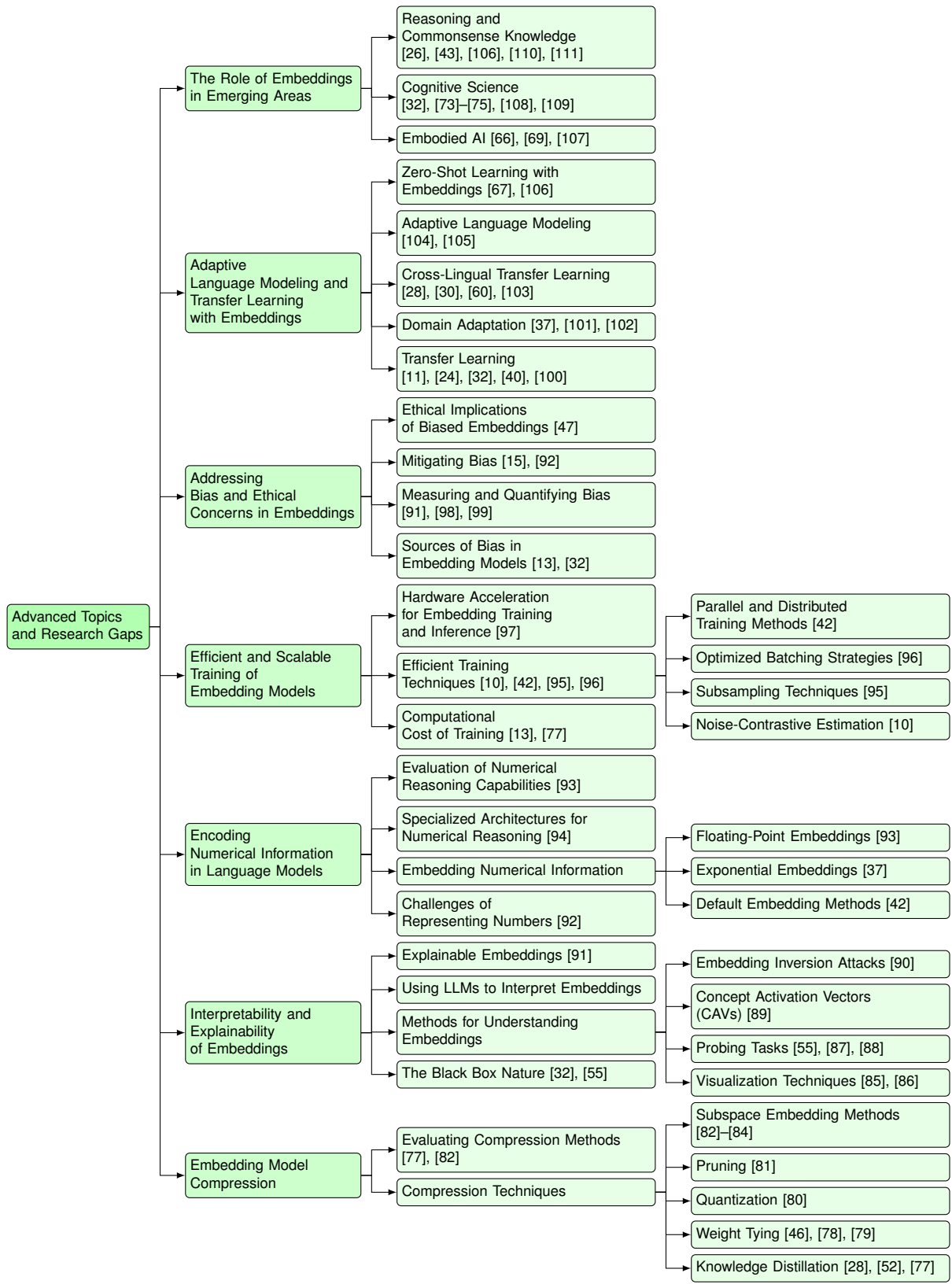

\subsection{The Role of Embeddings in Emerging Areas}

\subsubsection{Embodied AI: Grounding Language in Sensorimotor Experiences}
Embodied AI seeks to ground language understanding in sensorimotor experiences, enabling agents to interact with and reason about the physical world. Multimodal embeddings are crucial for bridging linguistic instructions and robotic actions. \cite{jaeyong_deep_2015} introduced an algorithm that learns a shared embedding space for point-cloud data, natural language, and manipulation trajectories, enhancing accuracy and inference in robotic tasks. This approach allows robots to generalize from past interactions with objects to manipulate new objects based on linguistic commands. Further advancing embodied AI, \cite{driess_palm-e_2023} proposed PaLM-E, a multimodal language model that integrates continuous sensor data (e.g., visual, state estimation) into a large language model, excelling in embodied tasks like robotic manipulation planning, visual question answering, and captioning. Kosmos-2 also contributes to this field by grounding text in visual information, such as bounding boxes, which supports complex multimodal tasks like phrase grounding and referring expression comprehension \cite{peng_kosmos-2_2023}.

\subsubsection{Cognitive Science: Modeling Human Language Processing with Embeddings}
Embeddings have become a valuable tool in cognitive science for modeling human language processing and understanding the neural basis of language. Studies reveal a striking correspondence between brain activity and artificial embeddings. \cite{ariel_brain_2022} observed that brain embeddings in Broca's area share a similar geometry with contextual embeddings from deep language models, allowing cross-domain predictions and indicating a shared representational space. Further research, such as \cite{ariel_alignment_2024}, reinforces this shared geometry’s predictive power, suggesting that embedding models provide a computational framework for language representation in the brain. Additionally, \cite{andrea_family_2023} explored subject-specific semantic representations, finding that language models can capture individual differences in processing personally familiar entities. \cite{goldstein_correspondence_2022} extended these insights by highlighting a correspondence between deep language model (DLM) layers and the temporal dynamics of brain activity during language processing. These findings suggest that artificial embeddings offer valuable insights into neural mechanisms underlying language processing \cite{yossi_analysis_2017}.

\subsubsection{Reasoning and Commonsense Knowledge: Integrating Embeddings with Probabilistic Models and Knowledge Graphs}
Integrating embeddings with probabilistic models and knowledge graphs enhances commonsense reasoning and knowledge representation in language models. \cite{mirza_language_2022} introduced a model-independent loss function that incorporates pre-trained language models into knowledge graph embeddings, improving knowledge graph completion accuracy. Additionally, \cite{rajat_complementary_2020} demonstrated that joint modeling of knowledge graphs and entity prediction with language models enhances performance on both tasks. \cite{mourad_word_2022} provided a comprehensive review of word embeddings and pre-trained language models, emphasizing their role in advancing NLP tasks. \cite{nancy_rethinking_2020} proposed a paradigm that evaluates language models based on their capacity for direct linguistic reasoning, suggesting that embeddings function as implicit knowledge repositories. \cite{mourad_word_2022} also explored integrating pre-trained language models into sequence-to-sequence models, highlighting their effectiveness in tasks like machine translation and summarization. These studies illustrate the potential of combining embeddings with knowledge representation techniques to boost reasoning capabilities. Moreover, \cite{sanjeev_latent_2015} introduced a generative model for word embeddings, providing a theoretical foundation for nonlinear models and explaining how embeddings support analogical reasoning.
\noindent \textbf{Fig \ref{fig:taxonomy_advanced}} presents a taxonomy of advanced topics and research gaps covered in this section.

\section{Future Directions}

\subsection{Open Challenges and Future Research Directions}

The rapid development of large language model (LLM) embeddings has led to impressive advancements in natural language processing. However, several challenges and open questions remain, which limit the full realization of their potential. 

One significant challenge is efficiently handling extremely long documents. Current embedding models struggle with long sequences, often resorting to truncation or segmentation, which can lead to information loss and increased computational overhead. Developing scalable models capable of processing extended contexts without compromising performance is essential. For example, Jina Embeddings 2 \cite{ondrej_precomputed_2021} addresses this by accommodating up to 8192 tokens, illustrating the potential of extended context windows for representing lengthy documents more effectively.

Improving interpretability and explainability of embeddings is another critical area for future research. Although embeddings effectively capture semantic relationships, their abstract nature often makes it difficult to understand how specific representations are derived. This lack of transparency limits their applicability in domains requiring explainable AI. Methods that increase interpretability, such as transforming abstract vectors into understandable narratives using LLMs, as explored in \cite{tennenholtz_demystifying_2023}, are essential for building trust and enabling deeper analysis of model behavior. Further research into the theoretical foundations of embedding models, including their relation to human cognition, as studied in \cite{ariel_brain_2022}, can enhance our understanding of their internal structures. Examining properties encoded in sentence representations, as done in \cite{yossi_analysis_2017}, is also a promising step towards clarifying these complex representations.

Addressing bias and ethical concerns in embeddings is paramount. Language models often reflect biases present in their training data, potentially perpetuating harmful stereotypes and discriminatory outcomes. Developing robust methods for detecting, mitigating, and preventing bias in embeddings is essential to ensure fairness and responsible LLM use. Research like \cite{shiva_evaluating_2023} highlights significant biases related to sensitive attributes such as gender identity and sexual orientation, underscoring the need for improved bias mitigation techniques. Furthermore, ethical concerns surrounding data privacy and potential malicious applications of embedding inversion attacks, as demonstrated in \cite{li_sentence_2023}, call for further research and the development of safeguards.

Grounding language models in non-textual modalities offers a promising avenue for enhancing world understanding. Current models primarily rely on text, which may limit their ability to capture grounded, real-world semantics. Integrating non-textual modalities such as images \cite{ryan_unifying_2014}, or audio and sensor data \cite{jaeyong_deep_2015}, could provide richer representations connecting language to real-world experiences. Additionally, grounding models in structured knowledge bases, as explored in \cite{j_connecting_2013}, could enhance their reasoning abilities and enable more complex inferences.

Developing models that integrate knowledge and reasoning capabilities into embeddings remains a crucial research direction. Current embeddings primarily capture statistical word correlations, often lacking deeper understanding of underlying knowledge and logical relationships. Integrating knowledge graphs, as investigated in \cite{mirza_language_2022}, or designing models for reasoning in the linguistic domain, as proposed in \cite{nancy_semantically_2019}, could support more sophisticated reasoning in LLMs. This includes exploring logical expressions for word semantics, as proposed in \cite{bhattarai_tsetlin_2023}, and methods for encoding temporal information into model weights, as studied in \cite{nylund_time_2023}. Additionally, further exploration of embedding models’ theoretical foundations, especially their relationship to human cognition as examined in \cite{richard_low-dimensional_2021}, could inform the development of more cognitively plausible embeddings.

\subsection{Potential Impact and Broader Implications}

Advancements in embedding models for Large Language Models (LLMs) hold significant promise for transforming a wide range of applications and domains.

One major area of impact is improved natural language understanding and generation. Although current LLMs have shown remarkable capabilities, they still struggle with certain aspects of language, such as capturing nuanced semantics, handling ambiguity, and generating coherent, contextually appropriate text. Enhanced embedding models could address these limitations by providing richer, more nuanced representations, enabling LLMs to better grasp the meaning and intent behind text. This advancement could lead to more accurate and fluent machine translation systems \cite{will_bilingual_2013}, more effective summarization tools \cite{sergey_pre-trained_2019}, and more engaging, informative dialogue systems \cite{hamid_deep_2015}.

Another promising area is the development of more sophisticated, human-like conversational AI systems. Current conversational AI often struggles with maintaining context, understanding complex queries, and generating appropriate responses. Improved embeddings could enable these systems to better understand the subtleties of human language, resulting in more natural, engaging interactions with chatbots and virtual assistants \cite{japa_question_2020}, more effective language tutoring systems \cite{victor_language_2016-1}, and adaptive language learning platforms \cite{trevor_cross-lingual_2017}.

Embedding models also have the potential to significantly enhance information retrieval and knowledge management systems. Traditional retrieval methods rely on keyword matching or basic similarity measures, which are limited when handling complex or ambiguous queries. Advanced embeddings could provide a more nuanced and semantically rich representation of documents and queries, leading to more accurate and relevant search results. This development could benefit search engines \cite{debasis_word_2015}, intelligent knowledge bases \cite{j_connecting_2013}, and sophisticated recommendation systems \cite{doddapaneni_user_2024}.

Beyond traditional NLP, embedding models are also gaining traction in emerging fields like robotics, embodied AI, and cognitive science. In robotics, embedding models can ground language in perception and action, enabling robots to understand and respond more effectively to human instructions \cite{jaeyong_deep_2015}. In embodied AI, embeddings facilitate the development of human-like agents that interact with the world in a more natural and intuitive way \cite{nancy_semantically_2019}. In cognitive science, embeddings offer insights into how the human brain represents and processes language, providing valuable understanding of the neural basis of language and cognition \cite{ariel_brain_2022}.

Lastly, advancements in embedding models could enable more personalized and adaptive language technologies. Personalized embeddings capture individual differences in language use and preferences \cite{charles_exploring_2020}, facilitating tailored language learning experiences. Adaptive language models that adjust based on user context can offer more relevant, useful information and services \cite{nikolaos_grounded_2020}. By incorporating user-specific information, these technologies can enhance the accessibility and usability of language systems for diverse audiences.

Moreover, embedding advancements could address broader societal challenges such as misinformation, bias, and accessibility. By capturing the nuances of language and concept relationships, embeddings can help detect and mitigate biases in language models \cite{shiva_evaluating_2023}, identify and counter misinformation \cite{linda_effect_2019}, and improve accessibility for people with disabilities. These advancements hold potential to foster more equitable and inclusive language technologies, benefiting society as a whole.

\bibliographystyle{ieeetr}  
\bibliography{references} 


\end{document}